\documentclass[conference]{IEEEtran}
\IEEEoverridecommandlockouts
\usepackage{cite}
\usepackage{amsmath,amssymb,amsfonts}
\usepackage{algorithmic}
\usepackage{graphicx}
\usepackage{textcomp}
\usepackage{xcolor}

\usepackage{amssymb}
\usepackage{CJKutf8}
\usepackage{float}
\usepackage{caption}
\usepackage{enumitem}
\setlist[itemize]{leftmargin=*}
\usepackage{amsmath}
\usepackage{url}
\usepackage{tikz}
\usetikzlibrary{positioning}

\usepackage{array}
\usepackage{booktabs} 
\usepackage{multirow}

\usepackage{graphicx}
\usepackage{tikz}
\usepackage{forest}
\usetikzlibrary{trees,positioning,shapes,shadows,arrows.meta}

\usepackage{amsthm}
\newtheorem{remark}{Remark}
\newtheorem{definition}{Definition}

\def\BibTeX{{\rm B\kern-.05em{\sc i\kern-.025em b}\kern-.08em
    T\kern-.1667em\lower.7ex\hbox{E}\kern-.125emX}}
\begin{document}

\title{Data Augmentation Techniques for Chinese Disease Name Normalization}

\author{\IEEEauthorblockN{\small Wenqian Cui} 
\IEEEauthorblockA{\scriptsize \textit{Dept. Computer Science and Enginnering} \\
\textit{The Chinese University of Hong Kong}\\
Hong Kong, China \\
wenqian.cui@link.cuhk.edu.hk}
\and
\IEEEauthorblockN{\small Xiangling Fu} 
\IEEEauthorblockA{\scriptsize \textit{School of Computer Science} \\
\textit{(National Pilot Software Engineering School)} \\
\textit{Beijing University of Posts and Telecommunications}\\
Beijing, China \\
fuxiangling@bupt.edu.cn}
\and
\IEEEauthorblockN{\small Shaohui Liu} 
\IEEEauthorblockA{\scriptsize \textit{School of Computer Science} \\
\textit{(National Pilot Software Engineering School)} \\
\textit{Beijing University of Posts and Telecommunications}\\
Beijing, China \\
shaohuiliu@bupt.edu.cn}
\and
\IEEEauthorblockN{\small Mingjun Gu} 
\IEEEauthorblockA{\scriptsize \textit{School of Computer Science} \\
\textit{(National Pilot Software Engineering School)} \\
\textit{Beijing University of Posts and Telecommunications}\\
Beijing, China \\
belong2summer@gmail.com}
\and
\IEEEauthorblockN{\small Xien Liu} 
\IEEEauthorblockA{\scriptsize \textit{Dept. Electronic Engineering} \\
\textit{Tsinghua University}\\
Beijing, China \\
xeliu@mail.tsinghua.edu.cn}
\and
\IEEEauthorblockN{\small Ji Wu}
\IEEEauthorblockA{\scriptsize \textit{Dept. Electronic Engineering} \\
\textit{College of AI}\\
\textit{Tsinghua University}\\
Beijing, China \\
wuji\_ee@mail.tsinghua.edu.cn}
\and
\IEEEauthorblockN{\small Irwin King}
\IEEEauthorblockA{\scriptsize \textit{Dept. Computer Science and Enginnering} \\
\textit{The Chinese University of Hong Kong} \\
Hong Kong, China \\
king@cse.cuhk.edu.hk}
}

\maketitle

\begin{abstract}
Disease name normalization is an important task in the medical domain. It classifies disease names written in various formats into standardized names, serving as a fundamental component in smart healthcare systems for various disease-related functions. Nevertheless, the most significant obstacle to existing disease name normalization systems is the severe shortage of training data. Consequently, we present a novel data augmentation approach that includes a series of data augmentation techniques and some supporting modules to help mitigate the problem. 
Through extensive experimentation, we illustrate that our proposed approach exhibits significant performance improvements across various baseline models and training objectives, particularly in scenarios with limited training data\footnote{We recommend you to read the full version of this paper at https://arxiv.org/abs/2306.01931}.
\end{abstract}

\begin{IEEEkeywords}
Data Augmentation, Disease Name Normalization, Medical Natural Language Processing
\end{IEEEkeywords}

\section{Introduction}

Disease names play a pivotal role in modern intelligent healthcare systems as it is involved in diverse tasks such as intelligent consultation \cite{chipcdn_kdd}, auxiliary diagnosis \cite{auxiliary_diagnosis1,auxiliary_diagnosis2}, automated International Classification of Diseases (ICD) coding \cite{automatic_icd_coding1, automatic_icd_coding2, automatic_icd_coding3}, Diagnosis-Related Groups prediction \cite{drg1, drg2}, etc. However, in clinical settings, doctors often write disease names according to their own habits and preferences, leading to numerous variations for the same disease. Therefore, to carry out additional operations on disease names, it is necessary to normalize them into standard names. As a result, disease name normalization, which entails classifying the diagnosis terms in clinical documents to standard names or classifications, plays a critical role in the ecosystem.

One of the main challenges in the disease normalization task is data scarcity. Specifically, a substantial proportion of disease names and concepts are typically not covered in the training set, leading to few-shot or zero-shot scenarios in the normalization process. For example, in CHIP-CDN dataset \cite{zhang2021cblue}, only about 25\% of all the diseases are provided. In this case, it is extremely difficult for the models to gain comprehensive knowledge about the disease system.
Although collecting more data seems to be a natural solution to address this challenge, it is more difficult to perform in the medical field due to privacy concerns and the requirement for expertise. 
Hence, in this work, we utilize data augmentation as a workaround to address the data scarcity problem.



We design a novel data augmentation approach including a set of data augmentation methods and some supporting modules for Chinese disease name normalization tasks called Disease Data Augmentation (DDA). Our data augmentation methods are designed to provide the models with an extensive understanding of disease names, particularly those that are absent in the original training set.
Our experiments demonstrate that our DDA approach outperforms all other data augmentation counterparts and effectively enhances the performance of various disease name normalization baselines. Furthermore, our approach can perform much better with smaller datasets and can achieve nearly 80\% of the full performance even when no data from the training set is provided.

\begin{figure*}[t]
    \centering
    \includegraphics[width=0.8\textwidth]{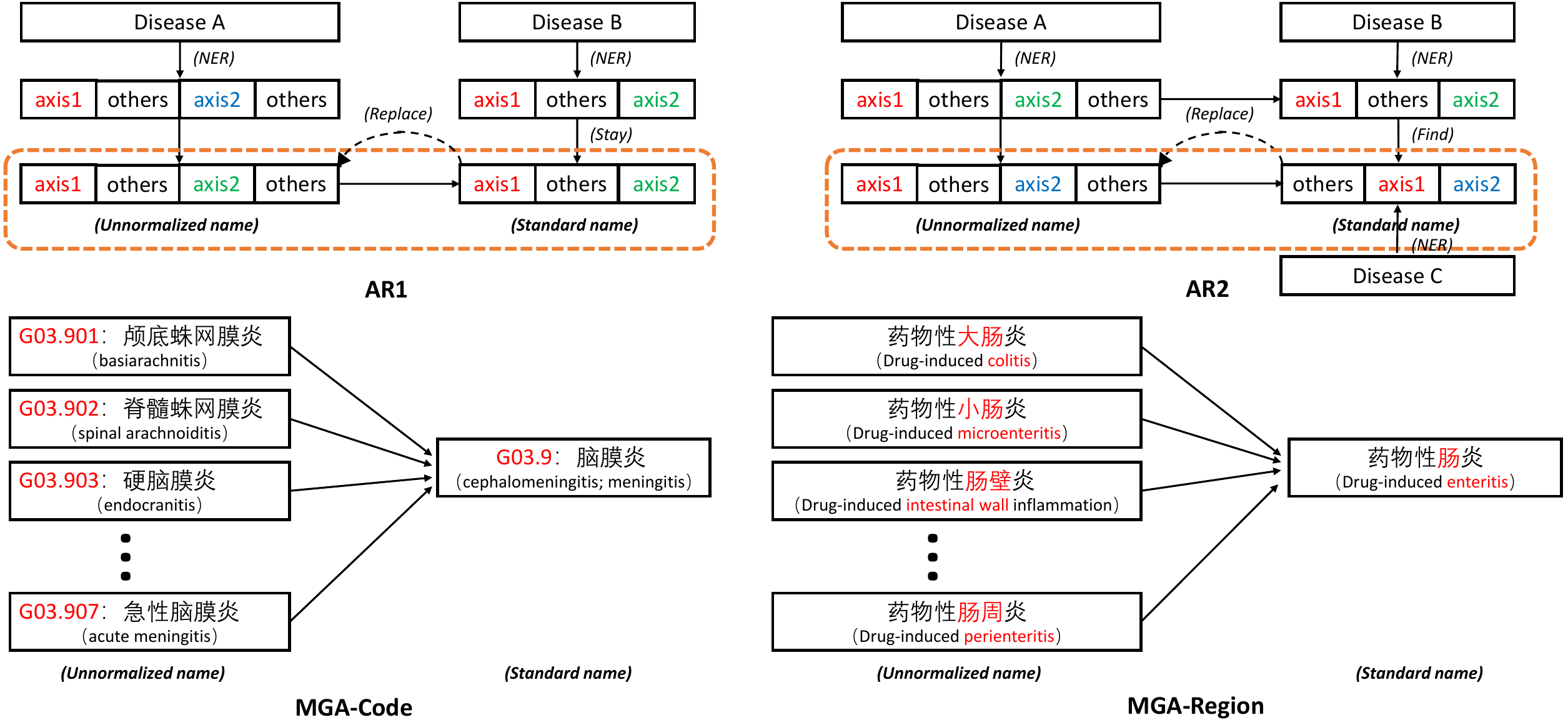}
    \caption{Illustration of our proposed data augmentation techniques. The upper portion of the figure depicts the Axis-word Replacement methods, and the lower portion depicts the Multi-Granularity Aggregation methods.}
    \label{methods}
\end{figure*}

\section{Proposed Methodology}
This section introduces the task definition of disease name normalization and our proposed data augmentation approach.

\begin{definition}[Disease Name Normalization]
Let $D = \{d_1, d_2, ..., d_n\}$ be the set of extracted disease terms from clinical documents, and $S = \{s_1, s_2, ..., s_m\}$ be the set of standard disease names, we define disease name normalization as $N(d_i) = \arg\max_{s_j \in S} P(s_j | d_i)$.
\end{definition}

\begin{definition}[Axis Word]
Axis words are the word components within disease names. We define three main axis words used in our approach. Disease Center: The minimal term that describes the nature of a disease. Anatomical Region: A part of the human body that has actual meaning in anatomy. Disease Characteristic: The characteristic of a disease that indicates the subtype or the cause of the disease. Take ``\begin{CJK*}{UTF8}{gbsn}增生性毛发囊肿\end{CJK*}" (Proliferative Trichilemmal Cyst) as an example, \begin{CJK*}{UTF8}{gbsn}囊肿\end{CJK*} (Cyst), \begin{CJK*}{UTF8}{gbsn}毛发\end{CJK*} (Trichilemmal), and \begin{CJK*}{UTF8}{gbsn}增生性\end{CJK*} (Proliferative) are its Disease Center, Anatomical Region, and Disease Characteristic, respectively.
\end{definition}


\subsection{Named Entity Recognition Module}
The first module of our approach is a named entity recognition (NER) system to locate and identify the axis words from all the input disease names.
To build the NER system, we select 5,000 diseases from the International Classification of Diseases (ICD) system\footnote{We use ICD as the standard disease classification system.} \cite{icd10} and ask doctors to annotate the labels (i.e., the three axis words) in BIO format \cite{bio_tagging_ner}. 
We use the traditional ``BiLSTM + CRF" as the NER model architecture. Specifically, there are three BiLSTM layers \cite{lstm} with a hidden dimension of 100, a fully connected layer, and a CRF layer \cite{crf}. The model achieves a 0.794 micro F1 score in our final evaluation. 

\subsection{Data Augmentation Module}
The data augmentation module consists of four data augmentation methods, and they are divided into two main categories: Axis-word Replacement (AR) and Multi-Granularity Aggregation (MGA). The main purpose of our data augmentation methods is to provide the model with additional knowledge, so we focus on exploring the components and relationships within diseases to give the model a comprehensive understanding of the various components and the hierarchical classification system of disease names. Figure \ref{methods} illustrates the two categories and four types of data augmentation methods\footnote{We have open-sourced the augmentation code on GitHub at \url{https://github.com/dreamtheater123/disease_name_dataset}.}.

\subsubsection{Axis-word Replacement (AR)}
Axis-word Replacement method is designed based on the assumption that disease names exhibit \textbf{Structural Invariance} property. This means that replacing an axis word in a disease name with another word of the same type still results in a meaningful disease name. For example, when the anatomical region ``\begin{CJK*}{UTF8}{gbsn}髂\end{CJK*} (Iliac)" of the disease ``\begin{CJK*}{UTF8}{gbsn}髂总动脉夹层\end{CJK*} (Common iliac artery dissection)" is replaced by another region ``\begin{CJK*}{UTF8}{gbsn}颈\end{CJK*} (Carotid)", we derive a name with the same type of disease but locates in another region ``\begin{CJK*}{UTF8}{gbsn}颈总动脉夹层"\end{CJK*} (Common carotid artery dissection)". Since there are often matches of axis words between an unnormalized disease name and a standard disease name in the disease name normalization task, simultaneously replacing the same axis word in both the unnormalized name and the standard name can typically ensure that the newly generated pair will still match. We leverage both the ICD and task data (data from the disease name normalization training set) to perform Axis-word Replacement. The detailed descriptions of each category of Axis-word Replacements are as follows: 
\begin{itemize}
\item \textbf{AR1} (Figure \ref{methods}, top left corner): First, we select a pair of diseases (disease A and disease B) that share one or more axis words (axis1 in the figure) but differ in another axis word (axis2 in the figure). Then, we replace axis2 in disease A with the same axis2 in disease B.

\item \textbf{AR2} (Figure \ref{methods}, top right corner): First, we select a pair of unnormalized-standard diseases from the disease name normalization training set. Let the unnormalized disease be disease A, and the standard disease be disease B. Then, find disease C from the ICD system that shares one or more axis words (axis1 in the figure) but differs in another axis word (axis2). Finally, we replace axis2 in disease A to be the same axis2 in disease C, so that the replaced disease A and disease C can form a new disease name normalization pair.

\end{itemize}

\begin{remark}
For both AR1 and AR2, we can choose either of the three axis words to perform replacement.
\end{remark}


\subsubsection{Multi-Granularity Aggregation (MGA)}
Multi-Granularity Aggregation (MGA) method is designed based on the \textbf{hierarchical} structure of the ICD system. The granularity levels of this structure are organized by the length of the ICD codes. For example, in ICD-10 Beijing Clinical Version 601, the disease name of the 4-digit code ``A18.2" is ``\begin{CJK*}{UTF8}{gbsn}外周结核性淋巴结炎\end{CJK*} (Peripheral Tuberculous Lymphadenitis)", and it has in total 10 child diseases that have a fine-grained description, with 6-digit codes ranging from ``A18.201" to ``A18.210", such as ``A18.201: \begin{CJK*}{UTF8}{gbsn}腹股沟淋巴结结核\end{CJK*} (Inguinal lymph node tuberculosis)" and ``A18.202: \begin{CJK*}{UTF8}{gbsn}颌下淋巴结结核\end{CJK*} (Submandibular lymph node tuberculosis)". This shows that the ICD system exhibits a tree-like structure, where a coarse-defined disease can be associated with multiple fine-grained diseases. We implement MGA augmentation using the following methods:

\begin{itemize}
\item \textbf{MGA - Code}  (Figure \ref{methods}, bottom left corner): We assign the label of a 6-digit disease name to its corresponding 4-digit disease name. We refer to this method as ``aggregation" because typically a 4-digit disease name can be linked to several 6-digit disease names, allowing the model to learn which diseases are similar.

\item \textbf{MGA - Region}  (Figure \ref{methods}, bottom right corner): In addition to the ICD system, anatomical regions also exhibit a tree-like hierarchical structure, where smaller regions can be grouped together to form a larger region. We use an expert-annotated region tree to identify disease names that share the same center but where the region of one disease is the larger region of another. We then assign the classification labels of the smaller-region disease names to their corresponding larger-region disease names. 

\end{itemize}

\subsection{Semantic Filtering Module}
Since the augmented data might contain low-quality data, we design a filtering module to eliminate disease pairs with low confidence, based on the assumption that unnormalized names should closely resemble standard names. 
To measure the level of similarity, the first criterion is a normalized $n$-gram matching (ngm) score between an unnormalized disease name (UDN) and a standard disease name (SDN): 
\begin{equation}
\label{eq:n-gram}
\resizebox{0.44\textwidth}{!}{$
\text{ngm}(UDN, SDN) = \frac{{\sum_{n=1}^{\min(j, k)} \left| n\text{-gram}(UDN) \cap n\text{-gram}(SDN) \right|}}{{\min(j, k)}}
$}, 
\end{equation}
where $j$ and $k$ are the lengths of UDN and SDN, respectively. 
Specifically, for each pair, we generate $n$-grams from $n$ equals 1 to the length of the shorter name in the pair. We then calculate the number of matched pairs and divide it by the length of the shorter name. 
This equation measures the similarity in the character level.
The second criterion is a cosine similarity score between the contextual embeddings of UDN and SDN outputted by BERT \cite{devlin2018bert}, i.e., 
\begin{equation}
\label{eq:cosine_similarity}
\resizebox{0.44\textwidth}{!}{$
\text{{similarity}}(UDN, SDN) = \frac{{\text{{BERT}}(UDN) \cdot \text{{BERT}}(SDN)}}{{\|\text{{BERT}}(UDN)\| \|\text{{BERT}}(SDN)\|}}
$}, 
\end{equation}
which measures the similarity from the contextual semantic level.
The final dataset is derived by filtering out generated data pairs below the threshold of the normalized $n$-gram score or the cosine similarity score, 
\begin{equation}
\label{eq:overall_semantic_filtering}
\resizebox{0.44\textwidth}{!}{$
\begin{aligned}
\text{{Final Dataset}} = \{(GeneratedPairs) | & \text{{ngm}}(UDN, SDN) > \alpha \\
& \land \text{{similarity}}(UDN, SDN) > \beta\}, 
\end{aligned}
$}
\end{equation}
where we set $\alpha$ and $\beta$ to be the threshold for the normalized $n$-gram score and the cosine similarity score, respectively. In this work, we set $\alpha$ and $\beta$ to be 0.7 and 0.8, respectively. As a result, 419,472 pairs of disease names are generated.

\subsection{Training Paradigm}
We train the models in two steps: first, we use augmented data for pre-training, followed by original task data for fine-tuning. The reason is that although the semantic filtering module helps eliminate low-quality disease pairs from the augmented data, it can’t guarantee that all remaining names are genuine, which could negatively affect performance. Since our goal is to use a large volume of data to provide the model with extensive knowledge, we leverage generated disease pairs during pre-training. 

\section{Experiments}
This section contains the experimental results to show the effectiveness of our proposed DDA approach. We evaluate our DDA approach on CHIP-CDN, which is a Chinese disease name normalization dataset \cite{zhang2021cblue} containing 6,000, 2,000, and 1,000 unnormalized-standard disease pairs in training, validation, and test set, respectively. We assess our approach using four baseline models: BiLSTM \cite{lstm}, BERT-base \cite{devlin2018bert}, CDN-Baseline (from CBLUE) \cite{zhang2021cblue}, and B\textsc{i-hard}NCE \cite{chipcdn_kdd}.
For BiLSTM, we use two BiLSTM layers with a hidden dimension of 256 followed by an MLP layer for classification. For BERT-base, we employ the CLS vector \cite{devlin2018bert} for classification. The CDN-Baseline is based on the BERT-base model and follows a "recall-decide" approach, recalling all relevant disease names before making the final decision. B\textsc{i-hard}NCE is a contrastive learning-based method also based on BERT-base. We report all metrics on the validation set.

\begin{table*}[t]
\fontsize{9}{7}\selectfont
\centering
\caption{Top: comparison for the choice of different data augmentation approaches across multiple baseline models using the CHIP-CDN dataset. Middle: Ablation study for the DDA approach. Bottom: Comparison between the performance of zero-shot inference and full fine-tuning over various baseline models.}
\begin{tabular}{lccccc}
\toprule
DA Approaches & BiLSTM & BERT-base & CDN-Baseline & B\textsc{i-hard}NCE & B\textsc{i-hard}NCE \\
(Metric) & (Acc) & (Acc) & (F1) & (RECALL@5) & (NDCG@5) \\
\midrule
None & 0.455 & 0.558 & 0.554 & 0.857 & 0.816\\
EDA & 0.451 & 0.519 & 0.561 & 0.795 & 0.798\\
BT & 0.466 & 0.556 & 0.578 & 0.845 & 0.828\\
DDA (ours) & \textbf{0.518} & \textbf{0.579} & \textbf{0.592} & \textbf{0.866} & \textbf{0.840}\\
\midrule
DDA - AR & 0.487 & 0.568 & 0.588 & 0.861 & 0.833\\
DDA - MGA & 0.455 & 0.558 & 0.554 & 0.857 & 0.816\\
DDA - ngm & 0.505 & 0.572 & 0.581 & 0.858 & 0.830\\
DDA - similarity & 0.485 & 0.560 & 0.574 & 0.857 & 0.826\\
\midrule
Zero-Shot & 0.034 & 0.073 & 0.113 & 0.672 & 0.670\\
\bottomrule
\end{tabular}
\label{CDN_result}
\end{table*}

\begin{figure}[htbp]
    \centering
    \includegraphics[width=0.4\textwidth]{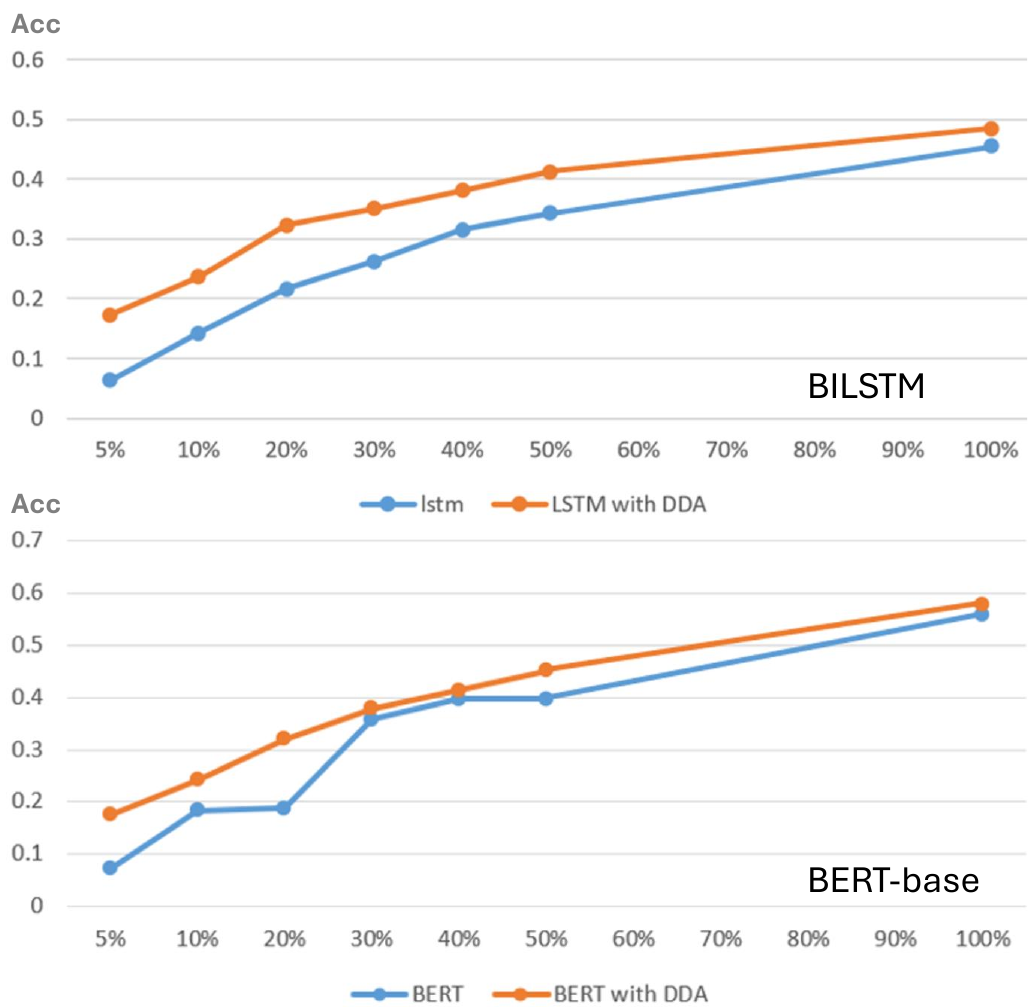}
    \caption{Performance comparison on smaller datasets for BiLSTM and BERT-base. The smaller datasets are derived by randomly sampling a portion of the CHIP-CDN training set. The validation set of CHIP-CDN stays the same.}
    \label{small_dataset}
\end{figure}

We first evaluate the effectiveness of our DDA approach by comparing it to two baseline approaches: EDA \cite{wei2019eda} and Back Translation (BT). As shown in the top part of Table \ref{CDN_result}, both EDA and back-translation have a detrimental impact on performance in certain scenarios (especially EDA), but DDA enhances performance across all scenarios. We then conduct an ablation study to illustrate the effectiveness of each category of data augmentation method in DDA. As shown in the middle part of Table \ref{CDN_result}, when removing either type of method or the semantic filtering rules one by one, we observe a decline in performance. This shows that all the data augmentation and filtering methods are effective.


We aim to evaluate performance improvements on smaller datasets from CHIP-CDN, as data scarcity is more pronounced in smaller datasets. We conduct experiments with training set sizes ranging from 5\% to 100\% of the original. As shown in Figure \ref{small_dataset}, the performance gap between whether to use our data augmentation or not is significantly larger when fewer training data is used. We further perform a zero-shot evaluation for all four pre-trained baseline models (without fine-tuning). B\textsc{i-hard}NCE is able to recover nearly 80\% of the full performance for RECALL@5 and NDCG@5, as depicted in the bottom part of Table \ref{CDN_result}.

\section{Conclusion}

In this work, we investigate disease name normalization in Chinese, highlighting the challenge of limited labeled data for model training. To address the challenge, we propose a novel data augmentation approach with two methods: Axis-word Replacement (AR) and Multi-Granularity Aggregation (MGA). These methods create new training pairs by manipulating disease name elements and aggregating based on the hierarchical structure of the ICD classification system. Our experiments show that this approach significantly improves performance across various baseline models compared to general text augmentation methods.

\section*{Acknowledgment}
The work described in this paper was partially supported by the Research Grants Council of the Hong Kong Special Administrative Region, China (CUHK 14222922, RGC GRF 2151185).



\bibliographystyle{IEEEtran}
\bibliography{references}


\end{document}